# Design and Implementation of a Fuzzy Adaptive Controller for Time-Varying Formation Leader-Follower Configuration of Nonholonomic Mobile Robots

Payam Nourizadeh[a], Aghil Yousefi-Koma[a,1], Moosa Ayati [b]

[a] *Centre of Advanced Systems & Technologies (CAST), School of Mechanical Engineering, College of Engineering, University of Tehran, Iran*
[b] *School of Mechanical Engineering, College of Engineering, University of Tehran, Iran*

*Abstract*— In this paper, a time-varying leader-follower formation control of nonholonomic mobile robots based on a trajectory tracking control strategy is considered. In the time-varying formation, the relative bearing and distance of each follower are variable parameters, and therefore, the followers can carry out various and complex behaviour even without changing the linear and angular velocities of the leader robot. After proposing the kinematic model of the time-varying leader-follower formation, the backstepping control method is exploited to keep the structure of the defined formation. The global stability of the formation is investigated using the Lyapunov theorem. Moreover, the designed nonlinear controller suffers from the ineffectual large input commands at the beginning of the formation. To rectify this problem, a fuzzy adaptive algorithm is proposed to improve the backstepping controller and the global stability of the resulting fuzzy adaptive backstepping controller is guaranteed. Considering the rate change of relative distance and bearing in the kinematic model of the leader-follower formation and controller design procedure, makes the formation more practical in dynamic and clutter environments, as well as capable of defining complicated behaviour for followers, and provides crash and obstacle avoidance without switching between different control strategies. Finally, the performance of the proposed kinematics model and designed controllers are investigated through simulations and experimental studies.

## I. INTRODUCTION

IN recent years, control strategies for a single wheeled mobile robot (WMR) have been developed due to the functionality of these robots in various environments and simple procedure of manufacturing WMRs [1]. Many researchers have worked on the modelling and control of a single WMR and the details have been published in a wide range of books and papers [2], [3]. However, as the defined manoeuvres for these robots became more complicated, it became necessary to utilize more than one robot. Using a group of mobile robots increases the robustness and flexibility of the system [4] and necessitates a system of cooperative control. In most cooperative control strategies, mobile robots should align in a defined formation to achieve the goal of the manoeuvre within different applications such as transportation and rescue operations [5]. In the literature, three basic formations for mobile robots have been defined: virtual structure, behaviour-based, and leader-follower [6]. Each of these formation strategies has their own advantages and disadvantages [7].

The leader-follower formation strategy consists of a leader robot that the other robots follow at a defined relative distance and bearing. The variety of potential applications for this formation has made the leader-follower formation popular, and its development is in continuous progress. In future, it will not be astonishing to see a group of autonomous vehicles in the fast lane of a highway traveling at the relative distance of less than 10 centimetres. The main advantage of this formation is gained by converting a formation control problem to a tracking control procedure in which the stability of the formation can be proved using mathematical theories. The main disadvantages of this formation are challenges such as the dependence of the formation on the performance of the leader robot, facilitating the operations of the leader robot independently to the followers, estimating linear and angular velocities [8], and localization of the formation [9]. As mentioned in [4], in the few studies on the formation control of multi-agent systems, it is the trajectory tracking problem that has been considered instead of the formation control problem. In contrast, the sensibility of the formation to the performance of the leader robot and the independent behaviour of the leader robot without feedback from the follower robots are the disadvantages of this formation; these can be rectified to make this formation more applicable. Furthermore, extracting the linear and angular velocity of the leader robot and the relative position of each follower are crucial to create this formation, a task that could be very challenging in practical experiments. In the case of a lack of data transmission between robots, the formation could fail the worst-case scenario. Nevertheless, this formation has been widely utilized in comparison with the other formation approaches.

[1] Corresponding author.
*E-mail addresses*: pnourizadeh@ut.ac.ir (P. Nourizadeh), aykoma@ut.ac.ir (A. Yousefi-Koma), m.ayati@ut.ac.ir (M. Ayati).

In the behaviour-based approach, the formation algorithm consists of a specific behaviour for every agent to keep the formation in place and to reach the system's defined goal while taking into consideration collision and obstacle avoidance [5], [10], [11], [12], [13]. The virtual structure formation strategy proposes a controlling algorithm to keep the agents in a defined rigid body. Papers [14] and [15] have contributed the main works on this strategy; see also these additional papers [7], [16], [17], [18], [19].

Das et al. [20] have supplied one of the primary studies on the leader-follower formation. They proposed a software program in order to control a group of nonholonomic WMRs using simple controllers. In that paper, the governing kinematic equations were derived and an input-output feedback linearization controller was designed to control the formation. Illustration of the local stability proof of the controller is the key point of the paper. In a study by Peng et al. [21], matrix and graph theory was utilized to control the formation of the nonholonomic WMRs and the stability of the controller was guaranteed using the Lyapunov theorem. Taking into consideration the information of the leader robot for only some of the followers is the main advantage of the approach examined in this paper. On the other hand, using the sign function in the controller input commands and consequently the high frequency chattering in the actuators is its drawback.

In the literature, the field of robust controllers [22], [23] and the problem of measuring the velocity of the leader robot [22], [24], [25] have been investigated. The work by Ghommam et al. [22] yields a leading study in this area. It considered robust formation control without measuring the velocity of the leader robot. They designed the controller based on a backstepping control algorithm, a neural network, and the Lyapunov stability theorem, and the architecture of the proposed controller was centralized. A backstepping controller algorithm is a well-known method for formation control of mobile robots. Chen et al. [26] proposed a trajectory tracking–based backstepping controller for car-like mobile robots and investigated the front-wheel driving and steering effect in the leader-follower formation via simulations and experimental studies. In a work by Peng et al. [4], a backstepping controller designed for this formation and rectified the velocity-jump problem in the input commands at the beginning of the formation using the bioinspired neurodynamic-based method. They illustrated the stability of the proposed controllers based on the Lyapunov approach. The results of the paper were demonstrated by simulation.

Other controlling methods have been investigated for this formation, such as feedback linearization [27], vision-based control [28], fuzzy and neuro-fuzzy theorem [29], [30], adaptive PID [31], and adaptive fuzzy sliding mode controllers [32]. Some researchers considered the issue of obstacle and crash avoidance [33], [34] and the effect of sliding and skidding in this formation [35].

In this paper, to develop the leader-follower formation, the element of time-varying formation is considered. Unfortunately, the problem of time-varying cooperative robots has not been considered widely in studies [36] and so far, to the best of our knowledge, the time-varying formation in the leader-follower approach has not been considered in the literature. In the conventional leader-follower formation, the relative positions of the followers are fixed and cannot change during the task. In the modelling and controller design process for this formation, the change and the rate of change of the relative distance and bearing have not been considered and as a result, this is a default constraint in the classic leader-follower architecture. The main advantages of the time-variable version of this formation are: (1) increased flexibility of the formation in cluttered and dynamic environments, (2) the ability to define complex behaviour for the followers, and (3) the possibility of obstacle and crash avoidance without switching between the controlling systems. The contribution of this paper is to propose a novel model and controller system to carry out the time-varying version of the leader-follower formation and to investigate the performance of the designed system in experimental studies.

At first, the kinematic model of a single robot is considered. Since the contribution of this paper is proposing a controller for the followers and also since the trajectory of the leader can be defined directly, controlling a single WMR has not been considered. Second, the kinematic model and error equation of the leader-follower formation based on the $l$-$\theta$ schema with consideration for the change rate of the relative distance and bearing of the followers are proposed. After deriving the kinematic model of the formation, the procedure of designing the backstepping controller based on the Lyapunov stability theorem is delivered. The designed backstepping controller like the other nonlinear controllers suffers from the high input commands at the beginning of the formation, where the amount of the tracking errors is large [4]. In order to rectify this velocity-jump problem, a fuzzy system is designed to adaptively tune the controller input commands' coefficients in a real-time situation.

Moreover, unlike most of the fuzzy controllers, the global stability of the proposed fuzzy adaptive backstepping controller is illustrated using the Lyapunov theorem, which makes the controller very practical and easy to implement in real experiments.

Finally, to illustrate the performance of the proposed kinematic model for time-varying leader-follower formation and both designed controllers, the results of the experiment and simulation studies are provided.

The paper is organized as follows: In section 2, the kinematic model of a single WMR and the trajectory tracking problem of the time-varying leader-follower formation are presented. Section 3 and 4, describes the backstepping and fuzzy adaptive backstepping controllers' design procedure. Section 5, illustrates and discusses the simulation and experimental data results. Finally, the conclusion of this research is given.

## II. MODELLING

### A. Kinematic Model

Figure 1, shows a schematic model of a single nonholonomic mobile robot. The rear wheels of the robot are active and the front wheel is the passive one. The planar postures of the robot at the centre of the rear axle are $p = [x, y, \theta]^T$.

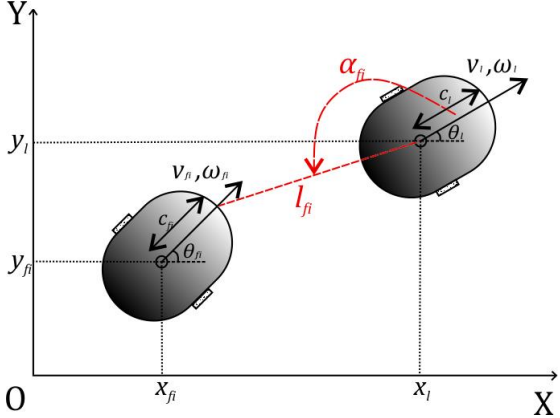

Figure 1. Geometry of a differential drive WMR and Leader-follower formation.

According to this model [2], the robot kinematic governing equation of motion is derived as

$$\begin{cases} \dot{x} = v \cos \theta - c\omega \sin \theta \\ \dot{y} = v \sin \theta + c\omega \cos \theta. \end{cases} \quad (1)$$
$$\dot{\theta} = \omega$$

In the equation above $v$ and $\omega$ are the linear and angular velocities of the robot, respectively. Also $c$ is the distance from the centre point of the rear axle to the front of the robot.

To develop a group of mobile robots and carry out the leader-follower formation, consider an arbitrary number ($m$+1) of the same nonholonomic mobile robots. The formation consists of a leader and $m$-mobile robots as followers (Figure 1). Each of the followers should track their desired trajectory which is defined based on the relative distance ($l_{fi}$) and the angle ($\alpha_{fi}$). The subscript $l$ and $fi$ refers to the leader and the $i^{th}$ follower robot, respectively. The desired trajectory of the $i^{th}$ follower robots is defined based on the geometrical condition of the formation such as

$$p_{fi}^d = [x_{fi}^d, y_{fi}^d, \theta_{fi}^d]^T =$$
$$\begin{bmatrix} x_{fi} - c\cos\theta_{fi} + l_{fi}^d \cos\beta_{fi} \\ y_{fi} - c\sin\theta_{fi} + l_{fi}^d \sin\beta_{fi} \\ \theta_{fi}^d \end{bmatrix}, \beta_{fi} = \alpha_{fi}^d + \theta_l. \quad (2)$$

Where in Eq. (2), superscript $d$ indicates the desired value of the parameter. The projectile Euclidean distance along the principal coordinates is derived in Eq. (3).

$$l_{fi} = \begin{bmatrix} l_{fi_x} \\ l_{fi_y} \end{bmatrix} = \begin{bmatrix} x_{fi} - x_l - c\cos\theta_l \\ y_{fi} - y_l - c\sin\theta_l \end{bmatrix} \quad (3)$$

In Eq.(3), $x_l$ and $y_l$ represent the coordination of the leader robot. In order to find the dynamic error equation of the formation, at first, the tracking error equation is defined as below.

$$e_{fi} = p_{fi}^d - p_{fi} = [e_{fi_x}, e_{fi_y}, e_{fi_\theta}]^T \quad (4)$$

where $p_{fi}$ is the actual position of the follower robot,

$$p_{fi} = [x_{fi}, y_{fi}, \theta_{fi}]^T. \quad (5)$$

The above error equation is defined in the XOY coordination system and in order to be represented in the follower robot local coordination system, the following mapping matrix is considered. Therefore, the error equation of the follower robot is determined using the transformation below.

$$\hat{e}_{fi} = [\hat{e}_{fi_x}, \hat{e}_{fi_y}, \hat{e}_{fi_\theta}]^T = \begin{bmatrix} \cos\theta_{fi} & \sin\theta_{fi} & 0 \\ -\sin\theta_{fi} & \cos\theta_{fi} & 0 \\ 0 & 0 & 1 \end{bmatrix} e_{fi} \quad (6)$$

Now, each of the matrix elements in Eq. (6) should be determined to obtain the rate of the error equation. For that reason, the following procedure is utilized for the local tracking error equation.

$$\begin{bmatrix} \hat{e}_{fi_x} \\ \hat{e}_{fi_y} \\ \hat{e}_{fi_\theta} \end{bmatrix} = \begin{bmatrix} \cos\theta_{fi}(x_{fi}^d - x_{fi}) + \sin\theta_{fi}(y_{fi}^d - y_{fi}) \\ -\sin\theta_{fi}(x_{fi}^d - x_{fi}) + \cos\theta_{fi}(y_{fi}^d - y_{fi}) \\ \theta_{fi}^d - \theta_{fi} \end{bmatrix} \quad (7)$$

The local orientation error is taken into account in the current format and the local direction errors are considered for the following procedure. Hence, taking the time derivative of the direction errors leads us to

$$\begin{bmatrix} \dot{\hat{e}}_{fi_x} \\ \dot{\hat{e}}_{fi_y} \end{bmatrix} =$$
$$\begin{bmatrix} \omega_{fi}\{-\sin\theta_{fi}(x_{fi}^d - x_{fi}) + \cos\theta_{fi}(y_{fi}^d - y_{fi})\} + \\ \cos\theta_{fi}(\dot{x}_{fi}^d - \dot{x}_{fi}) + \sin\theta_{fi}(\dot{y}_{fi}^d - \dot{y}_{fi}); \\ -\omega_{fi}\{\cos\theta_{fi}(x_{fi}^d - x_{fi}) + \sin\theta_{fi}(y_{fi}^d - y_{fi})\} - \\ \sin\theta_{fi}(\dot{x}_{fi}^d - \dot{x}_{fi}) + \cos\theta_{fi}(\dot{y}_{fi}^d - \dot{y}_{fi}); \end{bmatrix} \quad (8)$$

Substituting Eq. (2) into (8),

$$\begin{bmatrix} \dot{\hat{e}}_{fi_x} \\ \dot{\hat{e}}_{fi_y} \end{bmatrix}$$
$$= \begin{bmatrix} \omega_{fi}\hat{e}_{fi_y} - l_{fi}^d\omega_l\sin\gamma + \cos\theta_{fi}\dot{x}_{fi} + c\omega_l\cos\theta_{fi}\sin\theta_l + \\ \sin\theta_{fi}\dot{y}_{fi} - c\omega_l\sin\theta_{fi}\cos\theta_l - v_{fi} + l_{fi}^d\cos\gamma - \\ c\omega_l\sin\theta_{fi}\cos\theta_l - v_{fi} + l_{fi}^d\cos\gamma; \\ -\omega_{fi}\hat{e}_{fi_x} + l_{fi}^d\omega_l\cos\gamma - \sin\theta_{fi}\dot{x}_{fi} - c\omega_l\sin\theta_{fi}\sin\theta_l + \\ \cos\theta_{fi}\dot{y}_{fi} + \sin\theta_{fi}\dot{x}_{fi} - c\omega_l\cos\theta_{fi}\cos\theta_l - \\ \cos\theta_{fi}\dot{y}_{fi} + l_{fi}^d\sin\gamma - l_{fi}^d\dot{\alpha}_{fi}^d\cos\gamma; \end{bmatrix} \quad (9)$$

where;

$$\gamma = \alpha_{fi}^d + \theta_l - \theta_{fi} \quad (10)$$

In the equation above, two parameters $\dot{l}_{fi}^d$ and $\dot{\alpha}_{fi}^d$ have appeared. These parameters represent the time-varying leader-follower formation and have been considered in the dynamic error equation of follower motion. Substituting Eq. (1) into Eq. (9) leads to the dynamic error equation such as

$$\begin{bmatrix} \dot{\hat{e}}_{fi_x} \\ \dot{\hat{e}}_{fi_y} \end{bmatrix} = \quad (11)$$

$$\begin{bmatrix} \omega_{fi}\hat{e}_{fi_y} - l_{fi}^d\omega_l\sin\theta_{fi} + \cos\theta_{fi}\left(v_{fi}\cos\theta_{fi} - c\omega_{fi}\sin\theta_{fi}\right) + \\ c\omega_l\cos\theta_{fi}\sin\theta_l - c\omega_l\sin\theta_{fi}\cos\theta_l + \sin\theta_{fi} \times \\ \left(v_{fi}\sin\theta_{fi} + c\omega_{fi}\cos\theta_{fi}\right) - v_{fi} + \\ l_{fi}^d\cos\gamma - l_{fi}^d\dot{\alpha}_{fi}^d\sin\gamma\,; \\ -\omega_{fi}\hat{e}_{fi_x} + l_{fi}^d\omega_l\cos\gamma - \sin\theta_{fi}\left(v_{fi}\cos\theta_{fi} - c\omega_{fi}\sin\theta_{fi}\right) - \\ c\omega_l\sin\theta_{fi}\sin\theta_l + \cos\theta_{fi}\left(v_{fi}\sin\theta_{fi} + c\omega_{fi}\cos\theta_{fi}\right) - \\ c\omega_l\cos\theta_{fi}\cos\theta_l + \sin\theta_{fi}\left(v_{fi}\cos\theta_{fi} - c\omega_{fi}\sin\theta_{fi}\right) - \\ \cos\theta_{fi}\left(v_{fi}\sin\theta_{fi} + c\omega_{fi}\cos\theta_{fi}\right) + l_{fi}^d\sin\gamma - l_{fi}^d\dot{\alpha}_{fi}^d\cos\gamma\,; \end{bmatrix}.$$

Finally, after some simplification, the dynamic error is obtained in the following equation.

$$\begin{bmatrix} \dot{\hat{e}}_{fi_x} \\ \dot{\hat{e}}_{fi_y} \end{bmatrix} = \begin{bmatrix} \omega_{fi}\hat{e}_{fi_y} + v_l\cos\lambda + l_{fi}^d\cos\gamma - v_{fi} - \\ l_{fi}^d\omega_l\sin\gamma - l_{fi}^d\dot{\alpha}_{fi}^d\sin\gamma\,; \\ v_l\sin\lambda + l_{fi}^d\omega_l\cos\gamma + l_{fi}^d\sin\gamma + l_{fi}^d\dot{\alpha}_{fi}^d\cos\gamma - \\ \omega_{fi}\hat{e}_{fi_x} - c\omega_l\,; \end{bmatrix} \quad (12)$$

$\lambda = \theta_l - \theta_{fi}$

Therefore, the dynamic error can be written as the following equation.

$$\begin{bmatrix} \dot{\hat{e}}_{fi_x} \\ \dot{\hat{e}}_{fi_y} \\ \dot{\hat{e}}_{fi_\theta} \end{bmatrix}$$
$$= \begin{bmatrix} \omega_{fi}\hat{e}_{fi_y} + v_l\cos\lambda + l_{fi}^d\cos\gamma - v_{fi} - l_{fi}^d\omega_l\sin\gamma - \\ l_{fi}^d\dot{\alpha}_{fi}^d\sin\gamma\,; \\ v_l\sin\lambda + l_{fi}^d\omega_l\cos\gamma + l_{fi}^d\sin\gamma + l_{fi}^d\dot{\alpha}_{fi}^d\cos\gamma - \\ \omega_{fi}\hat{e}_{fi_x} - c\omega_l\,; \\ \omega_{fi}^d - \omega_{fi}\,; \end{bmatrix} \quad (13)$$

### B. Controller Design Procedure

Eq. (13) demonstrates the time derivative of the follower error in the leader-follower formation, and to track the leader robot, this error equation should approach zero. To make this happened, two control inputs ($v_{fi}, \omega_{fi}$) must be designed in a way that the stability of the closed-loop system is guaranteed. However, there is still an unknown parameter in the error equation of the follower robot, $\omega_{fi}^d$. Since these robots suffer from the nonholonomic constraint, this parameter cannot be chosen arbitrarily. Moreover, this desired angular velocity of the follower should be adjusted to guarantee the desired formation of the follower robots. The following designing procedure of the controller is based on the backstepping algorithm.

**Theorem 1**. *Consider $l_{fi}(t)$ and $\alpha_{fi}(t)$ to be the differentiable functions, therefore, there can be found the input linear and angular velocity for the follower robots that guarantees global asymptotical stability of the system in Eq. (13).*

**Proof**. First, the following Lyapunov candidate function is considered.

$$V_1(\hat{E}_1) = V_1\left(\hat{e}_{fi_x}, \hat{e}_{fi_y}\right) = (\hat{e}_{fi_x}^2 + \hat{e}_{fi_y}^2)/2 \quad (14)$$

The above Lyapunov function and its derivative functions are continuous and it has the following properties.

$$V_1\left(\hat{E}_1(0)\right) = 0, V_1(\hat{E}_1) > 0 \text{ where } \hat{E}_1 \neq 0 \quad (15)$$

Taking the first derivative of the Eq. (14) and substituting from Eq. (13) leads to,

$$\dot{V}_1(\hat{E}_1) = \hat{e}_{fi_x}\dot{\hat{e}}_{fi_x} + \hat{e}_{fi_y}\dot{\hat{e}}_{fi_y} = \left(\omega_{fi}\hat{e}_{fi_y} + v_l\cos\lambda + l_{fi}^d\cos\gamma - v_{fi} - l_{fi}^d\omega_l\sin\gamma - l_{fi}^d\dot{\alpha}_{fi}^d\sin\gamma\right)\hat{e}_{fi_x} + \left(v_l\sin\lambda + l_{fi}^d\omega_l\cos\gamma + l_{fi}^d\sin\gamma + l_{fi}^d\dot{\alpha}_{fi}^d\cos\gamma - \omega_{fi}\hat{e}_{fi_x} - c\omega_l\right)\hat{e}_{fi_y}. \quad (16)$$

According to Eq. (16), the input controllers should be chosen to obtain $\dot{V}_1(\hat{E}_1) \leq 0$. Therefore, choosing

$$v_{fi} = v_l\cos\lambda + l_{fi}^d\cos\gamma - l_{fi}^d\omega_l\sin\gamma + l_{fi}^d\dot{\alpha}_{fi}^d\sin\gamma + \kappa_1\hat{e}_{fi_x}$$
$$\omega_{fi} = \left(v_l\sin\lambda + l_{fi}^d\omega_l\cos\gamma + l_{fi}^d\sin\gamma + l_{fi}^d\dot{\alpha}_{fi}^d\cos\gamma + \kappa_2\hat{e}_{fi_y} + \kappa_3\hat{e}_{fi_\theta}\right)/c \quad (17)$$

leads the Eq. (16) to

$$\dot{V}_1(\hat{E}_1) = -\kappa_1\hat{e}_{fi_x}^2 - \kappa_2\hat{e}_{fi_y}^2 - \kappa_3\hat{e}_{fi_y}\hat{e}_{fi_\theta}. \quad (18)$$

In the equations above, $\kappa_i$ ($i = 1:3$) are positive constants. According to the equation above, since the sign of the derivative of $V_1(\hat{E}_1)$ is not semi-negative, the stability of the closed-loop system cannot be ensured. For that reason, the second Lyapunov candidate function is defined in the following equation and is based on $V_1(\hat{E}_1)$.

$$V_2(\hat{E}_2) = V_2\left(\hat{e}_{fi_x}, \hat{e}_{fi_y}, \hat{e}_{fi_\theta}\right) = V_1(\hat{E}_1) + \kappa_4\hat{e}_{fi_\theta}^2 \quad (19)$$

where $\kappa_4$ is a positive constant coefficient. The above Lyapunov function is positive ($V_2(\hat{E}_2) > 0$) and radially unbounded. The time derivative of the last Lyapunov function in Eq. (19) is determined such as:

$$\dot{V}_2(\hat{E}_2) = \dot{V}_1(\hat{E}_1) + 2\kappa_4\hat{e}_{fi_\theta}\dot{\hat{e}}_{fi_\theta}. \quad (20)$$

Substituting Eq. (18) and third row of Eq. (13) into (20) determines the time derivative of the Lyapunov function.

$$\dot{V}_2(\hat{E}_2) = -\kappa_1\hat{e}_{fi_x}^2 - \kappa_2\hat{e}_{fi_y}^2 - \kappa_3\hat{e}_{fi_y}\hat{e}_{fi_\theta} + 2\kappa_4(\omega_{fi}^d - \omega_{fi})\hat{e}_{fi_\theta} = -\kappa_1\hat{e}_{fi_x}^2 - \kappa_2\hat{e}_{fi_y}^2 - \kappa_3\hat{e}_{fi_y}\hat{e}_{fi_\theta} + 2\kappa_4[\omega_{fi}^d - (v_l\sin\lambda + l_{fi}^d\omega_l\cos\gamma + l_{fi}^d\sin\gamma + l_{fi}^d\dot{\alpha}_{fi}^d\cos\gamma + \kappa_2\hat{e}_{fi_y} + \kappa_3\hat{e}_{fi_\theta})/c]\hat{e}_{fi_\theta} \quad (21)$$

In the equation above, it is obvious that to determine the derivative of the Lyapunov function, the desired angular velocity should be defined. Moreover, even though considering any arbitrarily desired angular velocity does not demonstrate the instability of the closed-loop system, to guarantee the stability of the closed-loop system, defining the desired angular velocity is a critical selection. Therefore, the desired angular velocity of the follower is defined in the following equation to prove the stability of the follower robot.

$$\omega_{fi}^d = (v_l\sin\lambda + l_{fi}^d\omega_l\cos\gamma + l_{fi}^d\sin\gamma + l_{fi}^d\dot{\alpha}_{fi}^d\cos\gamma + 2\kappa_2\hat{e}_{fi_y})/c \quad (22)$$

Eq. (22) in (21) leads to,

$$\dot{V}_2(\hat{E}_2) = -\kappa_1\hat{e}_{fi_x}^2 - \kappa_2\hat{e}_{fi_y}^2 - \kappa_3\hat{e}_{fi_y}\hat{e}_{fi_\theta} + 2\kappa_4[(\kappa_2\hat{e}_{fi_y} - \kappa_3\hat{e}_{fi_\theta})/c]\hat{e}_{fi_\theta} \quad (23)$$

In Eq. (23) if $\kappa_4$ is defined as

$$\kappa_4 = \frac{c\kappa_3}{2\kappa_2}, \quad (24)$$

then

$$\dot{V}_2(\hat{E}_2) = -\kappa_1\hat{e}_{fi_x}^2 - \kappa_2\hat{e}_{fi_y}^2 - (\kappa_3^2/\kappa_2)\hat{e}_{fi_\theta}^2 = -\kappa_1\hat{e}_{fi_x}^2 - \kappa_2\hat{e}_{fi_y}^2 - \kappa_5\hat{e}_{fi_\theta}^2 \leq 0. \quad (25)$$

$\kappa_5 = \kappa_3^2/\kappa_2$.

According to Eq. (25), $\dot{V}_2(\tilde{E}_2) \leq 0$ and the equality is obtained only when $\tilde{E}_2 = g\left(\hat{e}_{fi_x}, \hat{e}_{fi_y}, \hat{e}_{fi_\theta}\right) = 0$. Therefore, the above equations demonstrate the local stability of the followers using the mentioned input controller. However, since the Lyapunov function in Eq. (19) is radially unbounded, the globally stability of the followers can be concluded. This means using the designed input controllers, guarantees the stability of the closed-loop system in global circumstances. ∎

Note that before starting the controller design procedure, the tracking error equation has been mapped using Eq. (6), and therefore, the calculated input velocity controller of the follower should be re-mapped again. For that reason, consider the following equation.

$$\begin{bmatrix} \dot{\hat{e}}_{fi_x} \\ \dot{\hat{e}}_{fi_y} \end{bmatrix}$$

$$= \begin{bmatrix} \omega_{fi}\left(-\sin\theta_{fi}\,e_{fi_x} + \cos\theta_{fi}\,e_{fi_y}\right) + \cos\theta_{fi}\left(\dot{x}_{fi}^d - \dot{x}_{fi}\right) + \\ \sin\theta_{fi}\left(\dot{y}_{fi}^d - \dot{y}_{fi}\right); \\ \omega_{fi}\left(-\cos\theta_{fi}\,e_{fi_x} + \sin\theta_{fi}\,e_{fi_y}\right) - \sin\theta_{fi}\left(\dot{x}_{fi}^d - \dot{x}_{fi}\right) + \\ \cos\theta_{fi}\left(\dot{y}_{fi}^d - \dot{y}_{fi}\right); \end{bmatrix} \quad (26)$$

Therefore,

$$\cos\theta_{fi}\,\dot{x}_{fi} + \sin\theta_{fi}\,\dot{y}_{fi} = \vartheta$$
$$\sin\theta_{fi}\,\dot{x}_{fi} - \cos\theta_{fi}\,\dot{y}_{fi} = \sigma. \quad (27)$$

Where in the equation above,

$$\vartheta = -\dot{\hat{e}}_{fi_x} - \omega_{fi}\sin\theta_{fi}\,e_{fi_x} + \\ \omega_{fi}\cos\theta_{fi}\,e_{fi_y} + \cos\theta_{fi}\,\dot{x}_{fi}^d + \sin\theta_{fi}\,\dot{y}_{fi}^d \quad (28)$$

and,

$$\sigma = \dot{\hat{e}}_{fi_y} + \omega_{fi}\cos\theta_{fi}\,e_{fi_x} + \\ \omega_{fi}\sin\theta_{fi}\,e_{fi_y} + \sin\theta_{fi}\,\dot{x}_{fi}^d - \cos\theta_{fi}\,\dot{y}_{fi}^d. \quad (29)$$

Finally, the next equation is used to find the explicit equation for $\dot{x}_{fi}$ and $\dot{y}_{fi}$.

$$\dot{x}_{fi} = (\cos\theta_{fi})\vartheta + (\sin\theta_{fi})\sigma$$
$$\dot{y}_{fi} = (\sin\theta_{fi})\vartheta - (\cos\theta_{fi})\sigma \quad (30)$$

*Remark 1.* Since in Eq. (6) $e_{fi_\theta} = \hat{e}_{fi_\theta}$, it is noticeable that the orientation and consequently the angular velocity of the follower robot has not been transformed.

## III. Fuzzy Adaptive Controller

In the last section, the backstepping controller to form the time-varying leader-follower formation has been designed and the global stability of the controller has been illustrated. The challenge of this nonlinear controller like other controllers is how to tune the constant in the input commands ($\kappa_i, i = 1:3$). Although considering positive values for these constants guarantees the global stability of the closed-loop system, a large amount of control input might be seen in the beginning of the formation. This phenomenon in controllers has been known as the velocity-jump problem, and to rectify this, various methods have been utilized. One of the fixing methods is changing the values of the coefficients ($\kappa_i, i = 1:3$) adaptively and since these variables multiply in the tracking errors and in the beginning of the motion where the amount of the tracking errors are large, in order to eliminate the velocity-jump problem, small positive values should be assigned to these variables during that moment of formation.

To adaptively tune these input variables, two fuzzy systems have been designed. Both of these fuzzy systems are based on the Proportional- Derivative (PD) controller and these systems have been defined to tune the first ($\kappa_1$) and the other coefficients ($\kappa_2, \kappa_3$) respectively. To consider the first fuzzy system, the $\hat{e}_{fi_x}$ and $\dot{\hat{e}}_{fi_\theta}$ have been considered as two inputs for this system and the output is $\kappa_1$. The rest of the coefficients should be extracted from the second fuzzy system. However, designing the Multi-Input Multi-Output (MIMO) fuzzy system is complicated and to avoid this problem, the two coefficients have been assumed equally ($\kappa_2 = \kappa_3$). This assumption leads us to the design of a Multi-Input Single-Output (MISO) fuzzy system, which is much faster and simpler than the MIMO one. Therefore, the inputs of this system are defined in the following equation and the output would be $\kappa_2$ or $\kappa_3$.

$$\tilde{e}_{y\theta} = \hat{e}_{fi_\theta} - \hat{e}_{fi_y}$$
$$\dot{\tilde{e}}_{y\theta} = \dot{\hat{e}}_{fi_\theta} - \dot{\hat{e}}_{fi_y} \quad (31)$$

For both systems, the membership function for inputs and output has been demonstrated in Figure 2 and Figure 3, respectively.

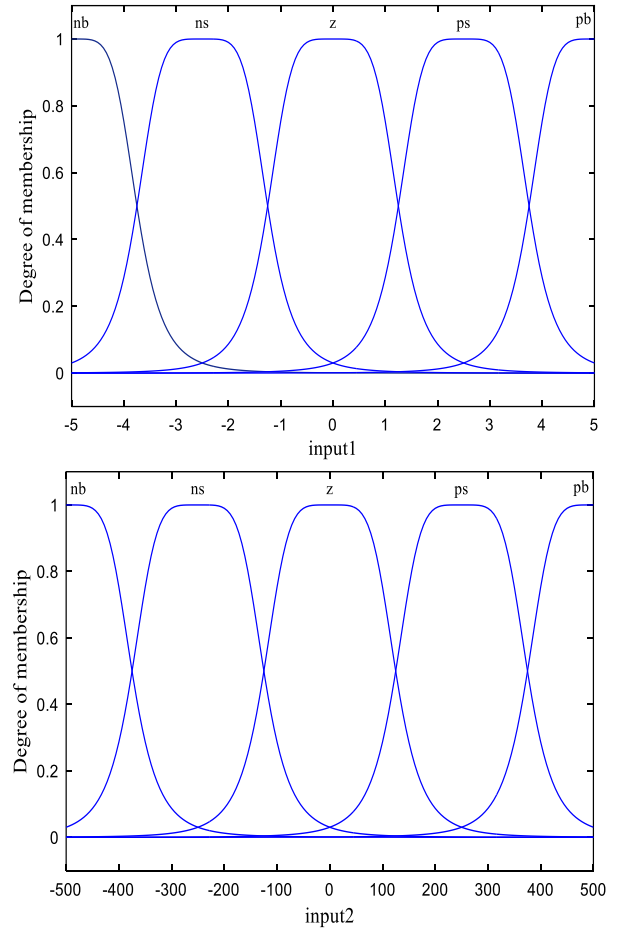

Figure 2. membership functions of the (up) $\hat{e}_{fi_x}$, $\tilde{e}_{y\theta}$ and (down) $\dot{\hat{e}}_{fi_\theta}$, $\dot{\hat{e}}_{y\theta}$.

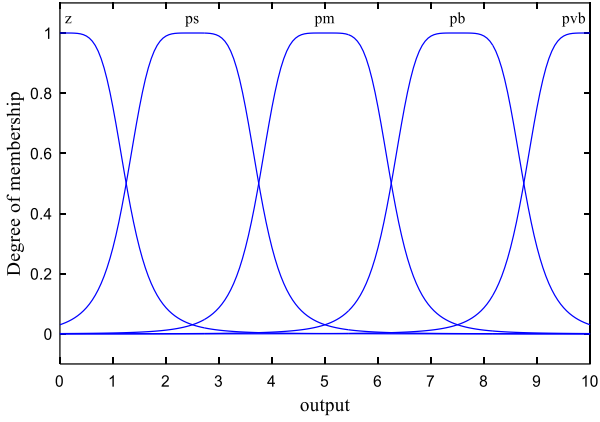

Figure 3. Membership functions of the output ($\kappa_i, i = 1:3$).

*Remark 2.* According to Eq. (25), the mentioned coefficients must be positive to guarantee the global stability of the closed-loop system. Therefore, the above membership functions have been defined in order to generate positive output.

In addition, the below fuzzy rule base has been defined to generate the mentioned coefficients (Table 1).

Table 1. Rules of the fuzzy system.

| | | | $e$ | | |
|---|---|---|---|---|---|
| | | $nb$ | $ns$ | $z$ | $ps$ | $pb$ |
| | $nb$ | $z$ | $z$ | $ps$ | $ps$ | $pm$ |
| | $ns$ | $z$ | $ps$ | $ps$ | $pm$ | $pm$ |
| $\dot{e}$ | $z$ | $z$ | $ps$ | $pm$ | $pb$ | $pb$ |
| | $ps$ | $z$ | $pm$ | $pb$ | $pb$ | $pvb$ |
| | $pb$ | $pm$ | $pb$ | $pb$ | $pvb$ | $pvb$ |

As is discussed in *Remark 2*, the output of the fuzzy system is positive to satisfy the stability constraint of the follower robot. Therefore, although in most fuzzy controllers the stability of the controller has not been proven, in the proposed controller, the global stability of the fuzzy adaptive controller has been illustrated using the Lyapunov theorem in Eq. (25).

## IV. RESULTS

In this section, the proposed controllers have been applied on two manufactured WMRs and the obtained results have been compared. There isn't any limitation for the number of follower robots, however, two similar robots have been considered, out of which one is the follower and the other is the leader robot.

### A. *Experimental setup*

Based on Figure 1, the WMR has two degrees of freedom and these two independent inputs move the robots. In order to carry out the leader-follower formation, two similar robots have been manufactured with two active wheels in the rear and one spherical passive wheel in the front. For each robot, two MX-106 Dynamixel actuators have been utilized. Its Maxon motor uses the PID controller and is capable of being used for both the position and velocity control problem with a very accurate contactless absolute encoder. Moreover, a VIVOTEK- IP8165HP camera has been employed on the top of the motion plane in the experimental setup to feedback the position of the robots. The camera detects the colourful stickers on the top of the robots and measures the position of the robots using real-time image processing techniques and two levels of noise filtering. The manufactured robots and the camera have been demonstrated in Figure 4.

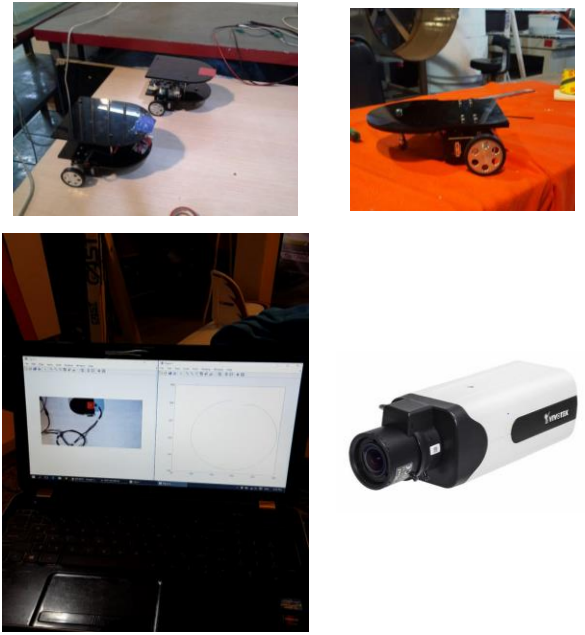

Figure 4. Experimental setup.

A desktop PC (Intel Core I7 CPU Q720 1.6 GHz, 64 bit, 4 GB RAM) with LAN connection to the IP camera has been utilized to control and carry out the image processing, and the controller has been implemented using MATLAB/Simulink software. The block-diagram of the closed-loop system has been illustrated in Figure 5. According to this figure, the desired path for the formation is obtained from the path planning algorithm. The nonlinear kinematic controller, as a high level controller, generates the input kinematic commands for the robots, which has been designed in sections 3 and 4. To complete the feedback control system, a localization algorithm is used. Moreover, the input torques of the 4 employed actuators are generated from the PID controller of the Dynamixel motors as a low level controller. In addition, the properties of the actuators have been indicated in Table 2. In order to illustrate the performance of the designed controllers, two different case studies have been defined. In the first case study, both of the results of the simulation and the experiment have been demonstrated and compared to show the functionality of the controllers in real implementation. Since the performance of the designed controller and the simulation results have been proved in the first case study, in the second case study a different manoeuvre has been defined and results have been demonstrated by simulation.

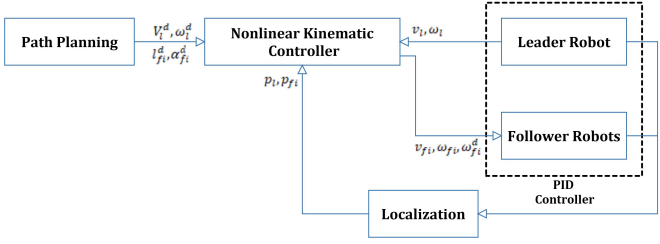

Figure 5. Architecture of the designed controller.

Table 2. Properties of the actuators.

| Property | Amount |
|---|---|
| Input voltage ($V$) | 12-14.3 |
| Input ampere ($A$) | 0.15-0.5 |
| P gain | 4 |
| I gain | 0.05 |
| D gain | 0.1 |

### 1) Case study A

The first configuration for this case study has been defined in order to illustrate the validity of the time-varying leader-follower kinematic model and the designed controller to carry out this formation. For this purpose and to concentrate on the effect of the proposed method on the follower robot, the trajectory of the leader robot has been considered as a straight line and therefore, the linear and angular velocity of this robot have been considered as 2 $cm/sec$ and zero, respectively. The initial conditions for the leader and follower robot have been considered as follows ($mm$, $rad$):

$$x_{0_l} = -50, y_{0_l} = 0, \theta_{0_l} = \pi/2$$
$$x_{0_f} = 400, y_{0_f} = -180, \theta_{0_f} = \pi/2$$
$$l_f^d = 80\sin(0.2t) + 300, l_f^d = 16\cos(0.2t) \tag{32}$$
$$\alpha_f^d = 3\pi/2, \dot{\alpha}_f^d = 0.$$

Where in the above equation, subscripts $l$ and $f$ indicate the leader and follower robot, respectively. According to Eq. (7), the initial formation errors are ($mm$, $rad$):

$$\hat{e}_{fxo} = 150, \hat{e}_{fyo} = 10, \hat{e}_{f_{\theta o}} = 0. \tag{33}$$

According to the designed backstepping controller in Eq. (17), the three positive constant inputs should be tuned ($\kappa_i, i = 1:3$), and therefore, 3, 3 and 4 have been assigned to these variables, respectively, using the trial and error method. The results of this case study have been illustrated in Figure 6. Also, in these pictures BC and FABC, respectively demonstrate the Backstepping Controller and Fuzzy Adaptive Backstepping Controller. In Figure 6, as was expected, the leader has been moved approximately on a straight line and since the tracking errors of the follower had the maximum amount at the first moment of the formation, this is when both input commands for the rear wheels had their maximum values (Figure 7 and Figure 8). The follower robot has been traversed along the desired trajectory using two different controllers (BC and FABC), and according to Figure 6, the follower robot has satisfied the defined constraints and has been kept its desired relative distance and bearing (both controllers). The results show that these controllers yield the time-varying leader-follower formation and the designed controllers make the closed-loop system stable, which has been verified using the

Lyapunov stability theorem. Except for the initial moment, the designed controllers have quite similar performance, however, this initial moment difference illustrates the advantage of the FABC. Although the trajectory of the follower is similar for both controllers, based on Figure 7 and Figure 8, the input controllers are quite different at the start of the formation. Figure 6 demonstrates that the follower robot has passed its trajectory with much less input control using FABC, and this controller has rectified the mentioned velocity-jump problem, which was the main aim of this controller. Table 3 provides more details about the input angular velocities for the following robot. Notice that both controllers have been designed to carry out the time-varying leader-follower formation and Figure 6 shows that the designed controllers satisfy the defined constraints. Moreover, although the following figures show that the BC is slightly faster in comparison with the FABC, the FABC generates a more smooth trajectory for the robot which is one of the advantages of fuzzy systems. Figure 9 illustrates the input adaptive gains, which are positive in order to make the closed-loop system stable.

Overall, the designed controllers accomplish all we set out to do. Both controllers generate the time-varying version of the leader-follower formation and these results verify the proposed equation of motion and the controllers. According to Figure 7 and Figure 8, the FABC carries out the mentioned formation with much lower input commands and rectifies the velocity-jump problem, which is very practical in experimental studies. Additionally, using less energy to carry out the formation, helps to save more power from the batteries and increases the duration of the missions. However, the BC generates the trajectory of the formation one minute faster than the other controller. Therefore, the mentioned advantages of the FABC in comparison with this small time difference, makes this controller more practical than the backstepping controller.

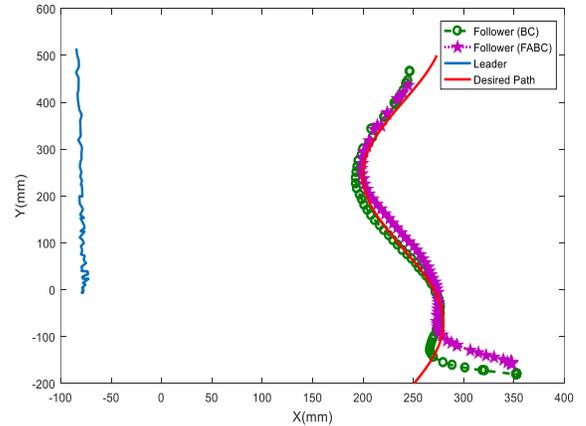

Figure 6. Trajectory of the leader and follower robots.

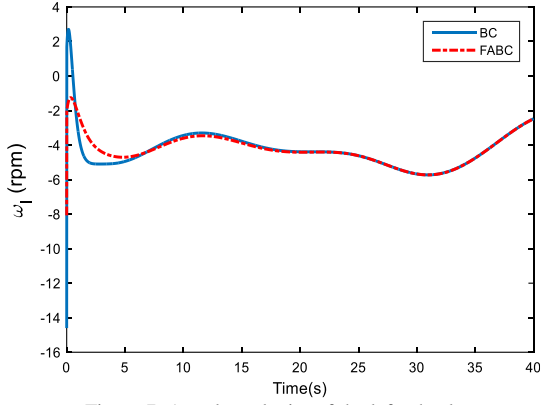

Figure 7. Angular velocity of the left wheel.

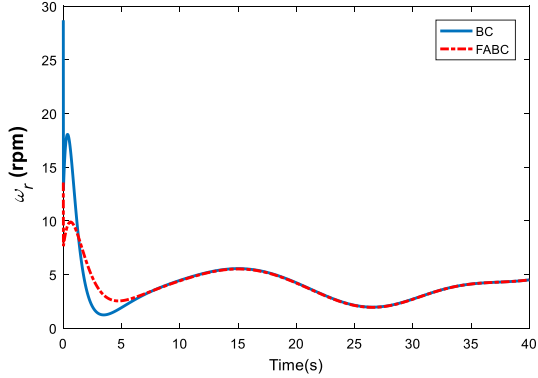

Figure 8. Angular velocity of the right wheel.

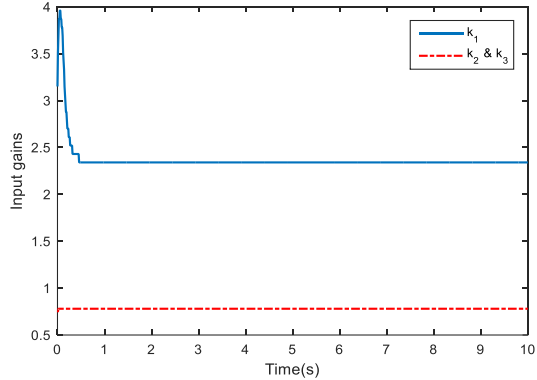

Figure 9. Input gains tuned by fuzzy system.

Table 3. Maximum input angular velocities ($rad/s$).

|  | BC | FABC | Percentage decrease (%) |
|---|---|---|---|
| Left wheel | -14.4 | -8.0 | 44 |
| Right wheel | 28 | 13.8 | 51 |

*2) Case study B*

The last case study verified the functionality of the designed controller and demonstrated the similarity of the results in theory and in experiments. In this section, another manoeuvre has been defined for the robots and results have been extracted from simulation. In this case study, a complex manoeuvre has been considered for the robots and both controllers have been examined through this simulation. This case study could show the flexibility of the designed formation and the capability of defining complicated trajectories for the followers. The initial conditions of this manoeuver have been mentioned below ($m$, $rad$) and the linear and angular velocity of the leader have been defined as 1 $m/sec$ and 0.5 $rad/sec$.

$$x_{0_l} = 0, y_{0_l} = 0, \theta_{0_l} = \pi/2$$
$$x_{0_f} = 1.5, y_{0_f} = -0.5, \theta_{0_f} = \pi/2$$
$$l_f^d = 1 - 2.15\cos(t)\cos(3t)\sin(t/5)$$
$$\dot{l}_f^d = 2.15[\sin(t)\cos(3t)\sin(t/5) \qquad (34)$$
$$+ 3\cos(t)\sin(3t)\sin(t/5)$$
$$- (1/5)\cos(t)\cos(3t)\cos(t/5)]$$
$$\alpha_f^d = 3\pi/2 \, , \dot{\alpha}_f^d = 0$$

Based on the conditions above, the initial formation errors have been determined as follows ($m$, $rad$):

$$\hat{e}_{f_{x0}} = 0.5, \hat{e}_{f_{y0}} = 0.5, \hat{e}_{f_{\theta0}} = 0. \qquad (35)$$

As has been mentioned, the linear and angular velocity of the leader are not equal and therefore, this robot will follow a non-circular trajectory. In Figure 10, the follower robot passes through a complex trajectory dissimilar to the conventional leader-follower formation. In the proposed formation method, the relative configuration of the follower changes with the defined rate (Figure 13) and as a result, the follower robot could be more functional which makes the formation more practical. This is the main different between the proposed method and the common leader-follower formation. As was noticed in case study A, the performances of the BC and FABC are quite similar and both controllers generate the time-varying formation. According to Figure 11, the FABC carries out the formation with less input energy, with a slight time difference; this behaviour is similar to those seen in the generated desired angular velocities, which are illustrated in Figure 14. Also, the tracking errors and adapted gains of the FABC have been shown in Figure 12.

*Remark 3. This case study illustrates the ability and flexibility of the followers to avoid possible obstacles in their working space.*

## V. CONCLUSION

In this paper a time-varying leader-follower formation of nonholonomic mobile robots has been investigated. To develop the time-varying formation, the changing rates of the relative distance and bearing have been studied in deriving the kinematic model of the formation and designing the nonlinear controllers. Initially, the backstepping technique has been utilized to design the controller to keep the formation in the defined condition and the global stability of the closed-loop system has been proved using the Lyapunov theorem. Simulations indicated that the designed backstepping controller -similar to other nonlinear controllers- suffered from the problematic high input commands when the tracking errors of the formation were large. In order to solve this problem, the fuzzy adaptive backstepping controller has been proposed and the Lyapunov theorem has illustrated the global stability of the designed controller. The experiment and simulation showed the performance of both designed controllers, and results indicated the fuzzy adaptive controller controls the formation with much lower input commands, which is very crucial during real experiments and helps to consume less energy from the power supply. Moreover, the results demonstrated that the time-varying leader-follower formation, in comparison with the conventional one, is more flexible, capable to implement in different environments and has the potential to avoid followers from collision with the obstacles and other agents.

## VI. ACKNOWLEDGMENTS

The authors would like to thank Iran National Science Foundation (INSF) for financial supporting this research.

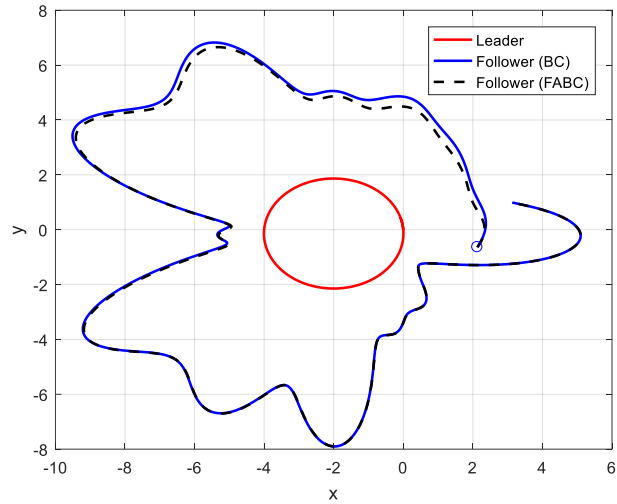

Figure 10. Trajectory of the robots in second maneuver (*m*).

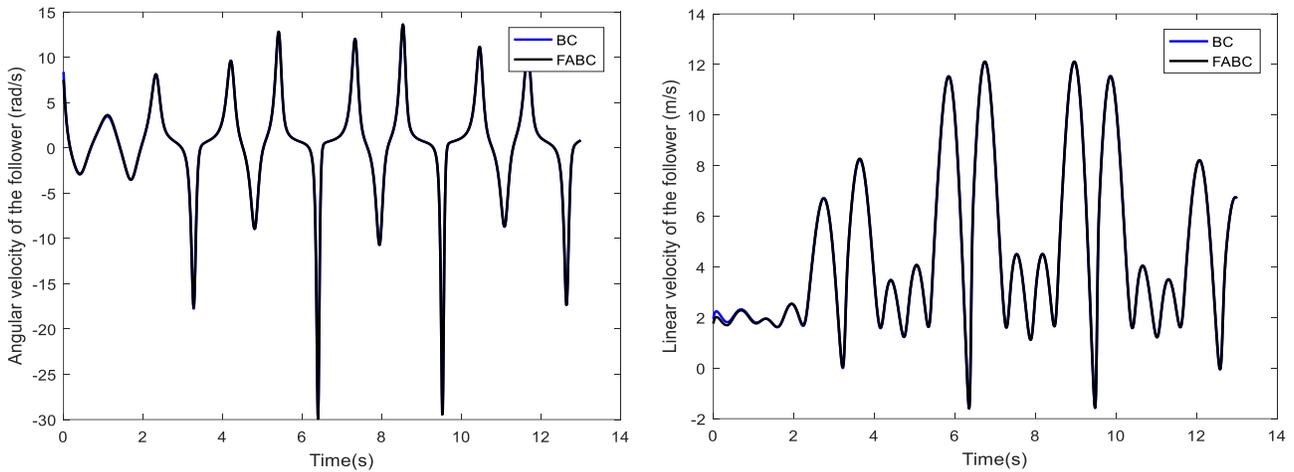

Figure 11. (Left) angular and (right) linear velocity of the follower.

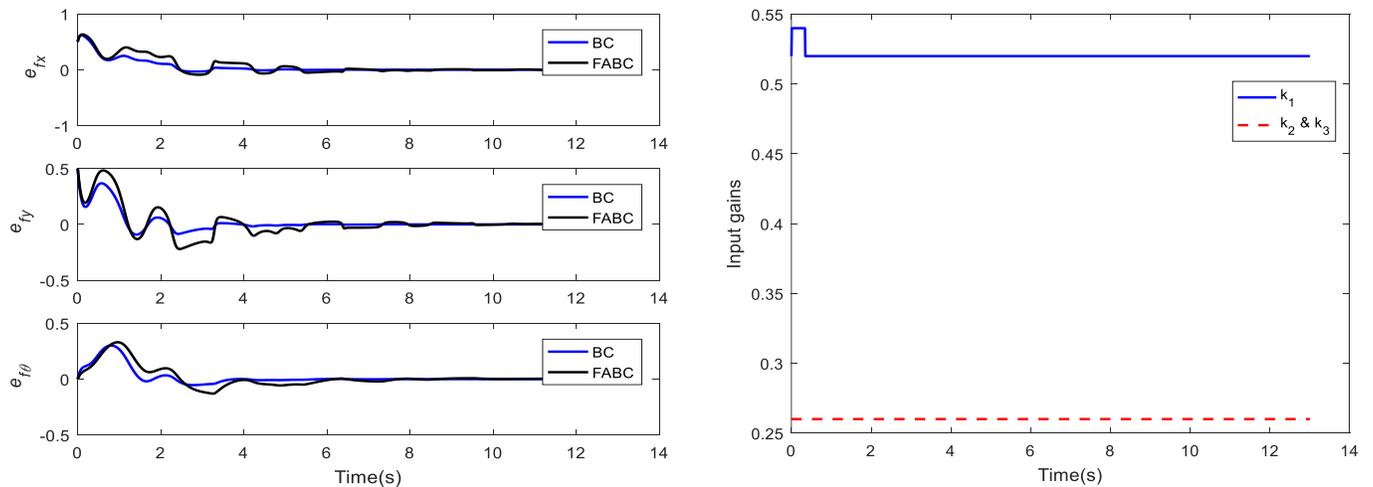

Figure 12. (Left) Tracking errors, (right) input gains tuned by fuzzy system.

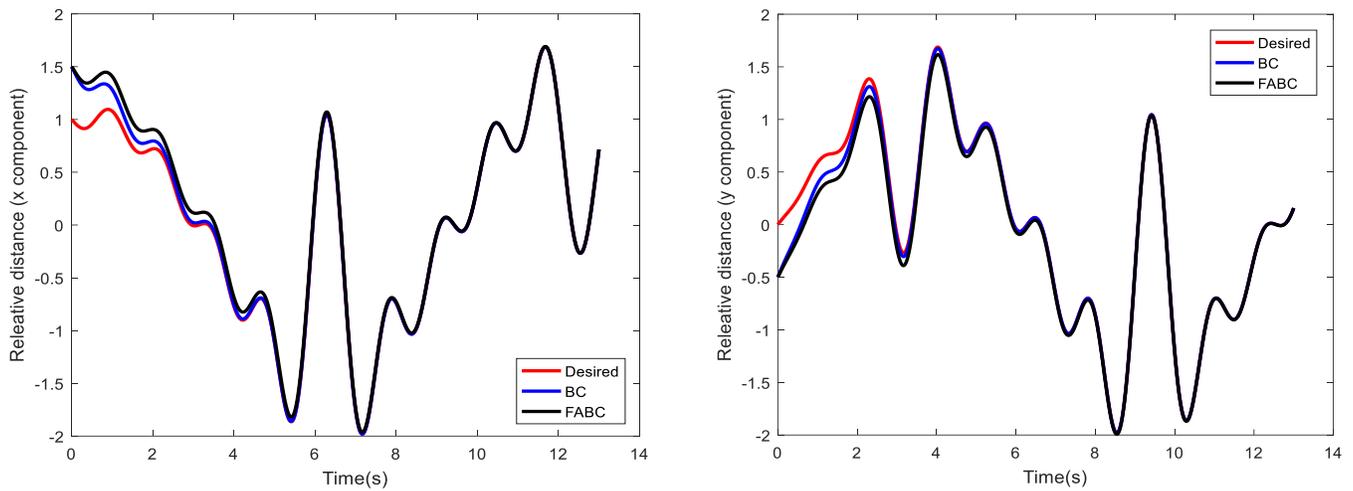

Figure 13. Relative distance (*m*).

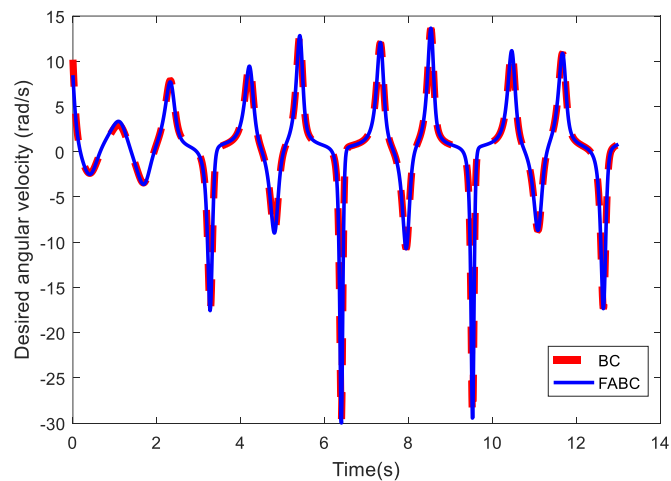

Figure 14. Desired angular velocity (*rad/s*).